\documentclass{bmvc2k}

\title{InSpaceType: Dataset and Benchmark for Reconsidering Cross-Space Type Performance in Indoor Monocular Depth}

\addauthor{Cho-Ying Wu}{choyingw@usc.edu}{1}
\addauthor{Quankai Gao}{quankaig@usc.edu}{1}
\addauthor{Chin-Cheng Hsu}{chincheh@usc.edu}{1}
\addauthor{Te-Lin Wu}{telinwu@g.ucla.edu}{2}
\addauthor{Jing-Wen Chen}{jchen885@usc.edu}{1}
\addauthor{Ulrich Neumann}{uneumann@usc.edu}{1}

\addinstitution{
 Computer Science Department\\
 University of Southern California\\
 California, USA
}
\addinstitution{
 Computer Science Department\\
 University of California, Los Angeles\\
 California, USA
}

\runninghead{Wu et al.}{InSpaceType}

\usepackage{graphicx}
\usepackage{epsfig}
\usepackage{amsmath}
\usepackage{amssymb}
\usepackage{url} 
\usepackage{booktabs}
\usepackage{amsfonts}       
\usepackage{nicefrac}       
\usepackage{microtype}      
\usepackage{array}
\usepackage{pifont}
\usepackage{capt-of,etoolbox}
\usepackage{multirow}
\usepackage{hhline}
\usepackage{bbding}
\usepackage{algorithm}
\usepackage{algorithmic}
\usepackage{boldline}
\usepackage{braket}
\usepackage{enumitem}
\usepackage{wrapfig,lipsum}
\usepackage{makecell}
\usepackage{colortbl}
\usepackage{times}
\usepackage{environ} 
\newcolumntype{?}{!{\vrule width 1.2pt}} 
\newcommand{\RNum}[1]{\uppercase\expandafter{\romannumeral #1\relax}}
\NewEnviron{NORMAL}{%
    \scalebox{0.88}{$\BODY$} 
} 
\usepackage{setspace}

\begin{document}

\maketitle

\begin{abstract}
\vspace{-0pt}
Indoor monocular depth estimation helps home automation, including robot navigation or AR/VR for surrounding perception. 
Most previous methods primarily experiment with the NYUv2 Dataset and concentrate on the overall performance in their evaluation. 
However, their robustness and generalization to diversely unseen types or categories for indoor spaces (spaces types) have yet to be discovered. Researchers may empirically find degraded performance in a released pretrained model on custom data or less-frequent types. 
This paper studies the common but easily overlooked factor- space type and realizes a model's performance variances across spaces. 
We present InSpaceType Dataset, a high-quality RGBD dataset for general indoor scenes, and benchmark 13 recent state-of-the-art methods on InSpaceType. 
Our examination shows that most of them suffer from performance imbalance between head and tailed types, and some top methods are even more severe. 
The work reveals and analyzes underlying bias in detail for transparency and robustness. 
We extend the analysis to a total of 4 datasets and discuss the best practice in synthetic data curation for training indoor monocular depth. Further, dataset ablation is conducted to find out the key factor in generalization.
This work marks the first in-depth investigation of performance variances across space types and, more importantly, releases useful tools, including datasets and codes, to closely examine your pretrained depth models. Data and code: \url{https://depthcomputation.github.io/DepthPublic/}
\vspace{-15pt}
\end{abstract}

\begin{figure}
    \centering
    \includegraphics[width=0.99\linewidth]{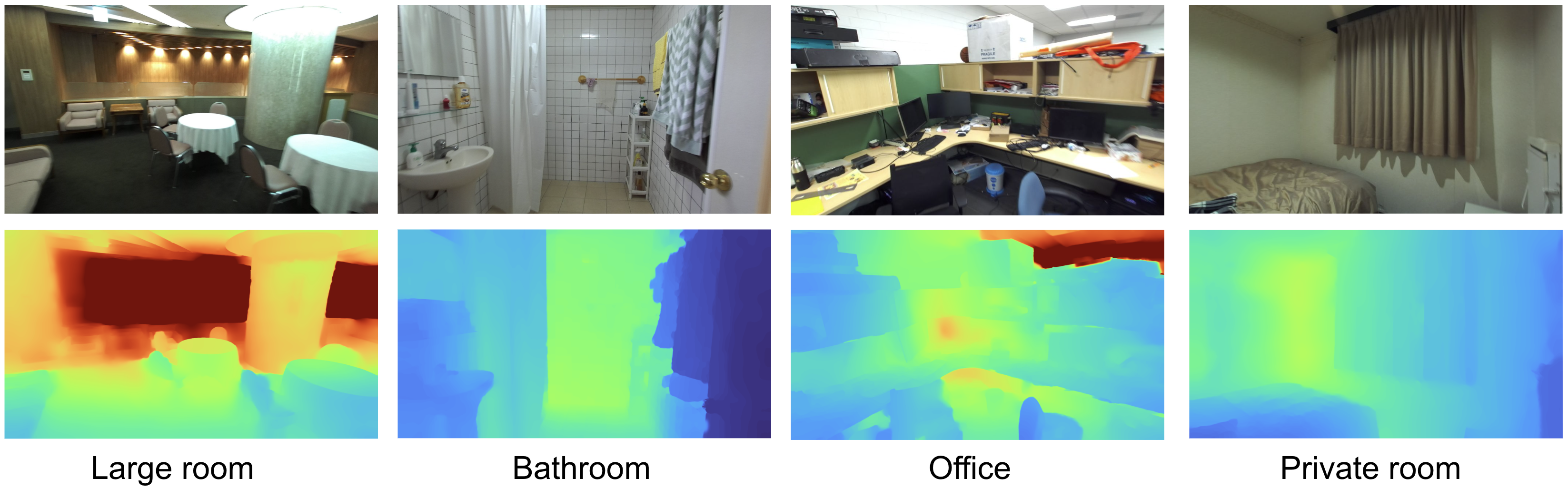}
    \includegraphics[width=0.99\linewidth]{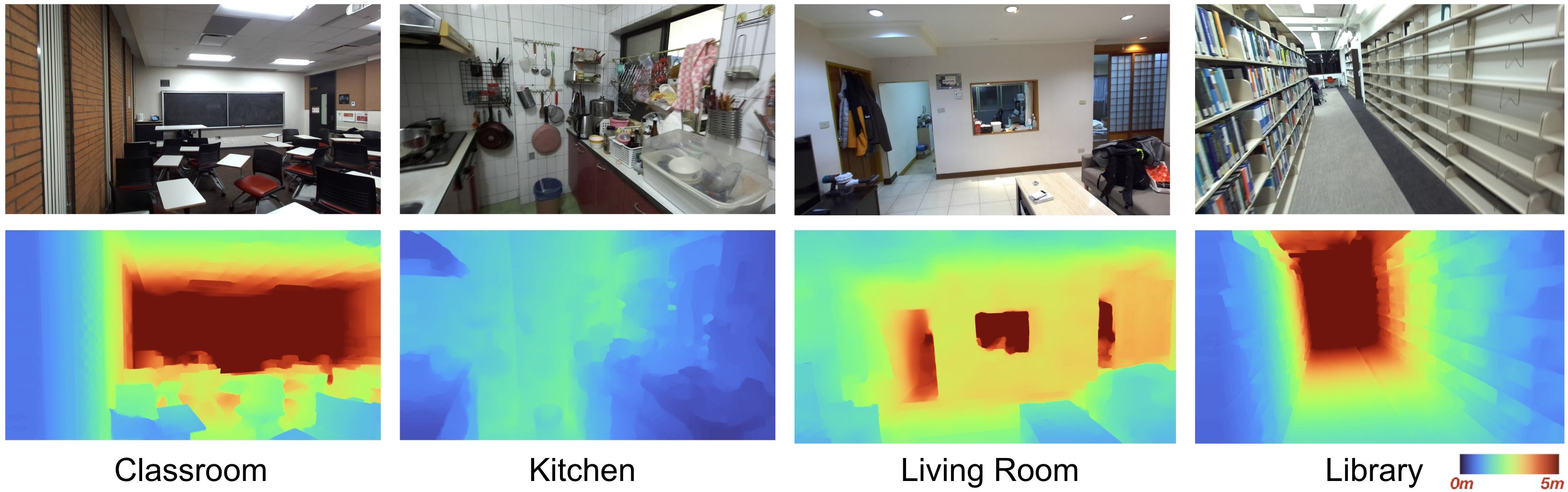}
    \vspace{-3pt}
    \caption{\textbf{InSpaceType data samples.} We color depth within 5 meters to show the near field. The captured data are dense, high-quality, and high-resolution for modern applications.}
    \vspace{-3pt}
    \label{fig_sample}
\end{figure}

\section{INTRODUCTION}
Depth estimation from monocular images is long-studied ~\cite{li2022depthformer,bian2021auto,jiang2021plnet,yuan2022new,kim2022global,bhat2022localbins,li2022binsformer,ramamonjisoa2021single,wu2020geometry,bhat2021adabins,fu2018deep,yin2019enforcing,liuva,liu2015deep,ricci2018monocular,yin2019enforcing} and widely used in industry for indoor automation nowadays, such as collision detection \cite{flacco2012depth} and navigation \cite{tai2018socially} for home robots, surrounding sensing in AR/VR~\cite{li2022real,diaz2017designing,mehringer2022stereopsis}, and RGBD approaches for biometrics surveillance~\cite{lee2020real}. It also helps learn better 3D representations in neural fields~\cite{deng2022depth}. 

Either recent works of using large models and training on multiple datasets such as ZoeDepth~\cite{bhat2023zoedepth}, DepthAnything~\cite{yang2024depth}, and Unidepth~\cite{piccinelli2024unidepth}, or prior methods of single-dataset train/ test~\cite{li2022depthformer,li2021structdepth,yuan2022new,kim2022global,bae2022irondepth,xie2023revealing,li2022binsformer,bhat2021adabins,bhat2022localbins,fu2018deep} primarily focus on the NYUv2 dataset~\cite{silberman2012indoor} for benchmark and evaluation. Though showing top performance on the benchmark, researchers may empirically find degraded performance on custom data or observe irregular depth prediction from unseen textures or
spaces that rarely appear in those models' training sets, such as oriental-style rooms. Robustness for real-world usage is not guaranteed.

We spot common shortcomings for the popular NYUv2 evaluation and benchmark: (1) Evaluation always reports metrics on the whole test set and overlooks the \textit{performance variances} across different space types\footnote{Space type in this work is defined by a space's function, such as bedroom, kitchen, or library. Some open-plan rooms, like condos, are included in the data, and we classify their types by the function of each captured scene}. A model may perform better in types appearing more often (head type) than those less-frequent types (tailed type). 
(2) NYUv2 is limited by a decade-old Kinect-v1 sensor that suffers from relatively low resolution (480$\times$640), noisy and blurry imaging, and limited measurable ranges. These downsides make the evaluation less reliable for modern robotics and high-resolution depth in AR/VR displays. 
Most prior depth estimation focuses on methodology and puts less focus on data and evaluation. 
However, as the monocular depth research matures, it is essential to investigate whether these models are ready for real usage with rigorous quality assessment.

We present a novel dataset, \textbf{InSpaceType}, to overcome the limitations and fulfill our study on performance variances across spaces. It is collected by a recent high-quality and high-resolution stereo camera with much better sensor characteristics than Kinect-v1.


\noindent\textbf{Space type-focused}. InSpaceType defines a hierarchical graph where each node is a type, and there are 26 leaf-node types in total that are representative and commonly seen in the real world. See the supplementary for the full graph. 13 recent high-performing methods with their pretrained models are used to benchmark the zero-shot performance for overall results and detailed space type breakdowns. 
Results show that imbalance exists in most methods, which are biased towards head types such as private room (MIM~\cite{xie2023revealing}'s $\delta_1$=94.54) and against tailed types such as large room (MIM's $\delta_1$=54.93). 

\noindent\textbf{Beyond NYUv2}. In addition to NYUv2, three other popular training sets are examined, including SimSIN \cite{wu2022toward}, 
UniSIN \cite{wu2022toward}, and Hypersim \cite{roberts2021hypersim}. 
We show space type breakdowns for models trained on them and reveal underlying bias to specify proper usage of them for future research.
In particular, we find synthetic datasets fail to reflect the real-scene complexity of cluttered and small objects, where models trained on synthetic datasets tend to underperform.


This work emphasizes the importance of the usually overlooked factor- \textit{space type}. In addition to analysis in the paper, we provide a quality assessment tool, including datasets and codes that show hierarchical type reports, to closely analyze your pretrained model.


\section{Related Work}



\subsection{Methods for Indoor Monocular Depth}
\label{related:method}

Several methods leverage paired image and groundtruth depth from indoor RGBD sensors to train depth estimation models. Many high-performing methods are transformer-based including NeWCRFs \cite{yuan2022new}, DepthFormer \cite{li2022depthformer}, PixelFormer \cite{agarwal2023attention}, GLPDepth \cite{kim2022global}, MIM \cite{xie2023revealing}, AiP-T \cite{Ning_2023_ICCV}. Some explore planarity or normal, including BTS \cite{lee2019big} and IronDepth \cite{bae2022irondepth}.
Some train on multiple datasets of indoor, outdoor, or unbounded scenes, including MiDaS \cite{Ranftl2020}, LeReS \cite{yin2021learning}, DPT \cite{Ranftl2021}, ZoeDepth \cite{bhat2023zoedepth}, VPD~\cite{zhao2023unleashing}, DepthAnything~\cite{yang2024depth}, and Unidepth~\cite{piccinelli2024unidepth}. These models learn generic image-to-depth mappings, but to estimate metric depth on a specific domain, most of them need a fine-tune set, usually NYUv2 for indoor scenes. 
Uniquely, Unidepth shows strong generalizability that does not need a fine-tune set and still slightly wins over DepthAnything on NYUv2.
There are few self-supervised methods that predict depth with physical indoor scales. For example, DistDepth \cite{wu2022toward} trains on simulated indoor stereo with left-right consistency, and GasMono \cite{Zhao_2023_ICCV} refines poses to recover scales.

Most of the above methods evaluate indoor performance only on NYUv2, and all of them only report overall performance, while performance variation across spaces is important for in-the-wild robustness.
This work pioneers the evaluation by space types. 
We show detailed prediction variances with transparency in robustness for these methods. 
Later results show that most top-performing methods still suffer from imbalances between head and tailed types, and even for some top methods, the gaps are prominent. 



\subsection{Evaluation Protocol for Indoor Monocular Depth} 
\label{dataset_proto}
Diode \cite{vasiljevic2019diode} collects both outdoor and indoor scenes with high-quality data but very low diversity of 2 scenes for evaluation only. 
IBims-1 \cite{koch2018evaluation} is limited to only 20 test scenes and 100 RGBD pairs.
VA \cite{wu2022toward} renders photo-realistic synthetic images but is limited to only one scene.
NYUv2~\cite{silberman2012indoor} is prevalent in indoor monocular depth and contains 654 RGBD test pairs. However, NYUv2 was collected by a decade-old device, Kinect-v1. First, its accurate and measurable ranges are only up to 3.5m from the hardware analysis~\cite{Kinect}, and farther measurements are highly noisy since the laser power dissipates quadratically with distance. Further, the sensor captures incomplete depth where lasers were absorbed, renders high imaging noise and blur by an older camera ISP, and outputs low-resolution RGBD data with only VGA size (640$\times$480). Compared with modern devices, NYUv2 is outdated in meeting the current application needs for robotics or high-resolution depth for AR/VR displays. 
NYUv2 contains small private rooms as the majority, and the benchmark only reports overall performance, which is thus dominated by the head type performance.

To overcome the limitations, we adopt a recently released high-quality stereo camera system, ZED-2i, to collect images and depth. From the sensor specification, ZED-2i estimates < 1\% error for up to 3m depth and < 5\% error for up to 15m depth. To verify, we use Kinect-v1 and ZED-2i to measure objects placed at 3m, 8m, and 15m. At 3m, both sensors predict about 0.8\% error. At 8m, ZED-2i's error is about 2.7\%, and Kinect's is about 10.8\%. At 15m, ZED-2i gets about 4.2\% error, but Kinect does not receive laser returns. Compared with Kinect, ZED-2i gets denser depth, much higher depth resolution (2208$\times$1242), and cleaner and sharper RGB imaging quality by a recent camera ISP.
Our collected InSpaceType covers general-purpose and highly diverse 88 indoor scenes, including private household spaces, workspace, campus scenes, and other functional spaces, amounting to 1260 RGBD pairs.
Our tool diagnoses a given model that shows the error and accuracy of each space type.



\section{InSpaceType Dataset}
\label{dataset}

\textbf{Device and Data Capture}.
The adopted stereo camera, Zed-2i, has a baseline of 12cm, a field of view (FOV) of $120^\circ$, a maximally measurable distance of 20m, and optimized dense depth maps by its on-device model with errors verified in Sec.~\ref{dataset_proto}.
Its setting gets 2208$\times$1242 resolution at 15 fps for aligned images and depth.
The device is anchored on a hand-held stabilizer for data capture.
Our environments cover household spaces, workspace, campus, and functional spaces with 26 leaf node types.
88 different environments are visited in total. 
Non-Lambertian surfaces (mirrors or highly reflective areas) are avoided.
We run at the full $120^\circ$ FOV and do not crop, zoom-in, or focus at flat walls that reduce cues for scene depth.
The camera moves at 6DoF with rotation and translation during capture.  
For the Euler rotation angles, yaw is unconstrained with $\pm 180^\circ$, where large yaw panning makes nearby frames less overlapped, pitch is within about $\pm 30^\circ$, and roll is within $\pm 10^\circ$. The constraints avoid reduced depth cues by larger pitch and roll, such as looking straight up to the ceiling.

\begin{figure}
    \centering
    \includegraphics[width=1.0\linewidth]{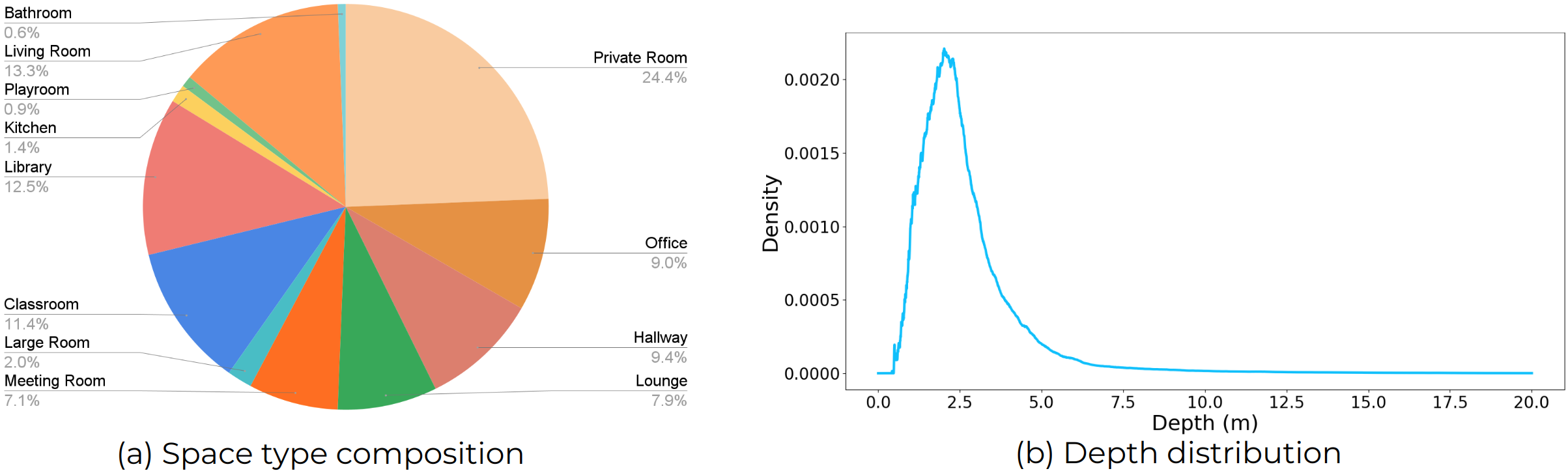}
    \vspace{-3pt}
    \caption{\textbf{Statistics of InSpaceType Evaluation Set.} Note that the type percentage affects the overall results in Table~\ref{table:2} but does not affect space type breakdowns in Table~\ref{table:25} and~\ref{table:3}, as each type performance is computed by \textit{averaging across all samples in a type} and compared to other types.}
    \vspace{-0pt}
    \label{fig_stat}
\end{figure}

\textbf{Evaluation set}.
We manually select 1260 images from all the environments to create the evaluation set.
Our selection criteria include (1) minimal motion blur, (2) not selecting from nearby 8 frames, and (3) sufficient depth cues (no single planes or zoom-in views) in the scenes.
Fig.~\ref{fig_sample} shows sample images, and Fig.~\ref{fig_stat} shows the dataset statistics. 
Note that each type is not evenly distributed, which reflects the natural distribution, such as libraries appearing less frequent than private rooms.
We do not favor or downsample common types to balance with the uncommon ones. 
To explain first, the distribution affects overall performance in Sec.~\ref{benchmark}-I, \textit{whose purpose is to know how a model performs naturally and does not favor uncommon types}. 
Space type breakdowns and analysis in Sec.~\ref{benchmark}-II and III cancel the uneven distribution
by averaging all samples in a type for evaluation metrics. \textit{It leaves out the factor of number of samples in a type} and compares them against the other types to reveal a model's strength and weakness.

\section{Benchmarks on InSpaceType}
\label{benchmark}
\textbf{[\RNum{1}: Overall performance]} The analysis aims to understand how a method performs naturally. We benchmark zero-shot cross-dataset performance on InSpaceType for 13 recent high-performing methods using their released indoor depth models, including DPT \cite{Ranftl2021}, GLPDepth \cite{kim2022global}, AdaBins \cite{bhat2021adabins}, PixelFormer \cite{agarwal2023attention}, NeWCRFs \cite{yuan2022new}, BTS \cite{lee2019big}, MIM \cite{xie2023revealing}, IronDepth \cite{bae2022irondepth}, Decomposition \cite{jun2022depth}, VPD \cite{zhao2023unleashing}, ZoeDepth \cite{bhat2023zoedepth}, DepthAnything~\cite{yang2024depth}, and Unidepth~\cite{piccinelli2024unidepth}. We adopt error (AbsRel, SqRel, RMSE; the lower, the better) and accuracy metrics ($\delta_1$, $\delta_2$, $\delta_3$ with a base factor of 1.25; the higher, the better) commonly used in monocular depth estimation. To compensate for different camera intrinsics between their training set and InSpaceType, we follow prior protocols for cross-dataset evaluation \cite{luo2020consistent, wu2023meta}: use median-scaling to align prediction and groundtruth's scales first and then calculate the metrics.

\begin{table}[tb!]
\centering
\caption{\textbf{InSpaceType benchmark: overall performance.} The best number is in bold, and the second-best is underlined. We include 13 recent high-performing methods.}
\vspace{-0pt}
\scriptsize
\label{table:2}
\begin{tabular}{|c|c|c|c|c|c|c|c|c|c|c|}
\hline
\cellcolor[HTML]{99CCFF} Method & \cellcolor[HTML]{99CCFF} Year &\cellcolor[HTML]{99CCFF} Architecture  & \cellcolor[HTML]{FAE5D3} AbsRel & \cellcolor[HTML]{FAE5D3} SqRel & \cellcolor[HTML]{FAE5D3} RMSE & \cellcolor[HTML]{D5F5E3} $\delta_1$ & \cellcolor[HTML]{D5F5E3} $\delta_2$ & \cellcolor[HTML]{D5F5E3} $\delta_3$ \\
\hline
BTS \cite{lee2019big}& arXiv'19 & DenseNet-161  & 0.1445 & 0.1162 & 0.5222 & 81.65 & 95.57 & 98.54 \\
AdaBins \cite{bhat2021adabins}& CVPR'21 & Unet+AdaBins  & 0.1333 & 0.0957 & 0.4922 & 83.64 & 96.36 & 98.92 \\
DPT \cite{Ranftl2021} & ICCV'21 & DPT-Hybrid  & 0.1224 & 0.0773 & 0.4616 & 85.96 & 97.17 & 99.19 \\
GLPDepth \cite{kim2022global}& arXiv'22 & Mit-b4 & 0.1239 & 0.0788 & 0.4527 & 86.05 & 97.36 & 99.16 \\
IronDepth \cite{bae2022irondepth}& BMVC'22 & EfficientNet-B5  & 0.1276 & 0.1022 & 0.4894 & 85.30 & 96.37 & 98.84 \\
Decomposition \cite{jun2022depth} & ECCV'22 & EfficientNet-B5  & 0.1278 & 0.1025 & 0.4899 & 85.25 & 96.35 & 98.83 \\
NeWCRFs \cite{yuan2022new}& CVPR'22 & Swin-L & 0.1251 & 0.0823 & 0.4541 & 86.04 & 96.68 & 98.94 \\
PixelFormer \cite{agarwal2023attention} & WACV'23 & Swin-L & 0.1225 & 0.0761 & 0.4392 & 86.08 & 97.03 & 99.10 \\
MIM \cite{xie2023revealing}& CVPR'23 & SwinV2-L & 0.1100 & 0.0679 & 0.4244 & 88.58 & 97.59 & 99.28 \\
VPD \cite{zhao2023unleashing} & ICCV'23 & StableDiffusion & 0.0983 & 0.0661 & 0.4138 & 90.23 & 97.48 & 99.22 \\
ZoeDepth (NK) \cite{bhat2023zoedepth}& arXiv'23 & BeiT-L & 0.0969 & 0.0527 & 0.3834 & 90.76 & 98.19 & 99.50 \\
ZoeDepth (N) \cite{bhat2023zoedepth}& arXiv'23 & BeiT-L & 0.0956 & 0.0528 & 0.3887 & \underline{90.81} & \underline{98.22} & 99.52 \\
DepthAnything \cite{yang2024depth}& CVPR'24 & ViT-L & \underline{0.0928} & \underline{0.0506} & \underline{0.3806} & 90.01 & 98.09 & \underline{99.54} \\
UniDepth \cite{piccinelli2024unidepth}& CVPR'24 & ViT-L & \textbf{0.0718} & \textbf{0.0349} & \textbf{0.3132} & \textbf{94.76} & \textbf{98.75} & \textbf{99.62} \\
\hline
\end{tabular}
\vspace{-4pt}
\end{table}

\begin{table*}[bt!]
\centering
\vspace{9pt}
\caption{\textbf{Space type breakdowns.} In each block of the table, we show a performance breakdown and an easy/hard type summary (See Sec.\ref{benchmark}-II). The first N-only block includes top methods, MIM and PixelFormer, for training on RGBD data only from NYUv2.
The M\&LS-Pre blocks show four top methods, ZoeDepth, VPD, DepthAnything, and Unidepth, for those pretrained on multiple datasets or using large-scale pretraining with/without fine-tuning. 
We highlight the best and worst scores of each metrics. For standard deviation in each block, where methods are in the same category and around the same publishing time, we show the lowest values in boldface, indicating low imbalance, and the second one with underlines.
}
\vspace{-0pt}
\tiny
\label{table:25}
\begin{tabular}{|c|c|c|c|c|c|c?c|c|c|c|c|c|}
\hline
 \cellcolor[HTML]{E9D66B} $\diamond$ \textbf{N-only} & \multicolumn{6}{c?}{\cellcolor[HTML]{E9D66B}MIM Results} & \multicolumn{6}{c|}{\cellcolor[HTML]{E9D66B}PixelFormer Results} \\

\cellcolor[HTML]{FCB381} Type Breakdown & \cellcolor[HTML]{FAE5D3} AbsRel & \cellcolor[HTML]{FAE5D3} SqRel & \cellcolor[HTML]{FAE5D3} RMSE & \cellcolor[HTML]{D5F5E3} $\delta_1$ & \cellcolor[HTML]{D5F5E3} $\delta_2$ & \cellcolor[HTML]{D5F5E3} $\delta_3$ & \cellcolor[HTML]{FAE5D3} AbsRel & \cellcolor[HTML]{FAE5D3} SqRel & \cellcolor[HTML]{FAE5D3} RMSE & \cellcolor[HTML]{D5F5E3} $\delta_1$ & \cellcolor[HTML]{D5F5E3} $\delta_2$ & \cellcolor[HTML]{D5F5E3} $\delta_3$ \\
\hline
Private room & 0.0927 & 0.0342 & 0.2556 & 92.05 & 98.99 & 99.83 & 0.1013 & \cellcolor[HTML]{A2D86C}0.0372 & 0.2638 & 90.17 & 98.70 & 99.80  \\
Office  & 0.1106 & 0.0532 & 0.3313  & 87.67 & 97.42 & 99.44 & 0.1271 & 0.0649 & 0.3658 & 84.60 & 96.38 & 99.19 \\ 
Hallway & 0.1229 & 0.0805 & 0.5463  & 85.66 & 96.52 & 98.89 & 0.1418 & 0.0880 & 0.5236 & 81.46 & 96.40 & 99.05 \\ 
Lounge & 0.1316 & 0.1290 & 0.7447 & 84.15 & 96.22 & 98.93 & 0.1500 & 0.1624 & 0.8215 & 79.92 & 94.79 & 98.41\\
Meeting room & 0.0984 & 0.0483 & 0.3636 & 91.70 & 98.69 & 99.64 & 0.1103 & 0.0551 & 0.3873 & 88.25 & 98.57 & 99.76\\ 
Large room & \cellcolor[HTML]{D8756C}0.2683 & \cellcolor[HTML]{D8756C}0.4499 & \cellcolor[HTML]{D8756C}1.3915 & \cellcolor[HTML]{D8756C}54.93 & \cellcolor[HTML]{D8756C}83.89 & \cellcolor[HTML]{D8756C}93.91 & \cellcolor[HTML]{D8756C}0.2125 & \cellcolor[HTML]{D8756C}0.3119 & \cellcolor[HTML]{D8756C}1.1396 & \cellcolor[HTML]{D8756C}67.71 & \cellcolor[HTML]{D8756C}88.59 & \cellcolor[HTML]{D8756C}95.40 \\ 
Classroom & 0.0781 & 0.0334 & 0.3071 & 94.52 & \cellcolor[HTML]{A2D86C}99.25 & \cellcolor[HTML]{A2D86C}99.83 & \cellcolor[HTML]{A2D86C}0.0912 & 0.0403 & 0.3368 & 91.37 & \cellcolor[HTML]{A2D86C}99.14 & \cellcolor[HTML]{A2D86C}99.86 \\
Library & 0.1342 & 0.0978 & 0.6281 & 85.61 & 96.57 & 98.57 & 0.1530 & 0.1242 & 0.6527 & 82.57 & 94.88 & 97.73\\
Kitchen & 0.1482 & 0.0791 & 0.3374 & 82.31 & 95.42 & 98.32 & 0.1819 & 0.0978 & 0.3521 & 78.86 & 93.05 & 96.63 \\ 
Playroom & \cellcolor[HTML]{A2D86C}0.0702 & \cellcolor[HTML]{A2D86C}0.0276 & \cellcolor[HTML]{A2D86C}0.2466 & \cellcolor[HTML]{A2D86C}94.54 & 98.25 & 99.79 & 0.1010 & 0.0476 & 0.3042 & \cellcolor[HTML]{A2D86C}92.11 & 97.23 & 99.15 \\ 
Living room & 0.1033 & 0.0502 & 0.3448 & 89.07 & 97.88 & 99.52 & 0.1132 & 0.0568 & 0.3601 & 87.35 & 97.31 & 99.34  \\
Bathroom & 0.1456 & 0.0772 & 0.2788 & 83.45 & 96.07 & 98.13 & 0.1439 & 0.0570 & \cellcolor[HTML]{A2D86C}0.2488 & 83.72 & 96.55 & 98.36 \\
\hlineB{2}
\textbf{Standard Deviation} & \underline{0.0395} & \underline{0.0559} & \underline{0.3384} & \underline{13.36} & \underline{7.50} & \underline{2.79} & \textbf{0.0361} & \textbf{0.0512} & \textbf{0.2972} & \textbf{6.78} & \textbf{3.58} & \textbf{1.93}\\
Value Max & 0.2683 & 0.4499 & 1.3915 & 94.54 & 99.25 & 99.83 & 0.2125 & 0.3119 & 1.1396 & 92.11 & 99.14 & 99.86 \\
Value Min & 0.0702 & 0.0276 & 0.2466 & 54.93 & 83.89 & 93.91 & 0.0912 & 0.0372 & 0.2488 & 67.71 & 88.59 & 95.40\\
\hline
\end{tabular}

\begin{tabular}{|l|l|l|}
\hline
\cellcolor[HTML]{FCB381} Easy/ Hard Summary & \multicolumn{1}{c|}{\cellcolor[HTML]{E9D66B}MIM} & \multicolumn{1}{c|}{\cellcolor[HTML]{E9D66B} PixelFormer} \\
\hline
\multicolumn{1}{|c|} {\cellcolor[HTML]{FAE5D3} \textbf{Easy type}} & \cellcolor[HTML]{FAE5D3} playroom, private room, classroom & \cellcolor[HTML]{FAE5D3} playroom, private room, classroom \\
\hline
\multicolumn{1}{|c|} {\cellcolor[HTML]{D5F5E3} \textbf{Hard type}} & \cellcolor[HTML]{D5F5E3} large room, lounge, library & \cellcolor[HTML]{D5F5E3} large room, lounge, library, hallway \\
\hline
\end{tabular}

\vspace{2pt}

\begin{tabular}{|c|c|c|c|c|c|c?c|c|c|c|c|c|}
\hline
\cellcolor[HTML]{99CCFF} $\dagger$ \textbf{M\&LS-Pre} & \multicolumn{6}{c?}{\cellcolor[HTML]{99CCFF}ZoeDepth(N) Results} & \multicolumn{6}{c|}{\cellcolor[HTML]{99CCFF}VPD Results}\\

\cellcolor[HTML]{FCB381} Type Breakdown & \cellcolor[HTML]{FAE5D3} AbsRel & \cellcolor[HTML]{FAE5D3} SqRel & \cellcolor[HTML]{FAE5D3} RMSE & \cellcolor[HTML]{D5F5E3} $\delta_1$ & \cellcolor[HTML]{D5F5E3} $\delta_2$ & \cellcolor[HTML]{D5F5E3} $\delta_3$ &\cellcolor[HTML]{FAE5D3} AbsRel & \cellcolor[HTML]{FAE5D3} SqRel & \cellcolor[HTML]{FAE5D3} RMSE & \cellcolor[HTML]{D5F5E3} $\delta_1$ & \cellcolor[HTML]{D5F5E3} $\delta_2$ & \cellcolor[HTML]{D5F5E3} $\delta_3$\\
\hline
Private room & 0.0798 & \cellcolor[HTML]{A2D86C}0.0253 & 0.2218 & 93.86 & \cellcolor[HTML]{A2D86C}99.44 & 99.90  & 0.0768 & 0.0325 & 0.2428 & 93.76 & 98.70 & 99.60 \\
Office  &  0.0978 & 0.0447 & 0.3102 & 90.07 & 97.80 & 99.52  & 0.1019 & 0.0548 & 0.3374 & 89.52 & 97.52 & 99.26\\ 
Hallway&  0.1193 & 0.0741 & 0.5271 & 85.84 & 96.95 & 99.10 & 0.1174 & 0.0828 & 0.5233  & 86.66 & 96.40 & 98.72\\ 
Lounge& 0.1172 & 0.1110 & 0.7033 & 86.61 & 97.02 & 99.11 & \cellcolor[HTML]{D8756C}0.1313 & 0.1286 & 0.7306 & 83.06 & 95.90 & \cellcolor[HTML]{A2D86C}98.83\\
Meeting room  &  0.0887 & 0.0391 & 0.3379 & 93.55 & 99.05 & 99.79  & 0.0992 & 0.0568 & 0.3742 & 91.51 & 98.26 & 99.43\\ 
Large room  &  \cellcolor[HTML]{D8756C}0.1565 & \cellcolor[HTML]{D8756C}0.1701 & \cellcolor[HTML]{D8756C}0.9157  & \cellcolor[HTML]{D8756C}77.19 & \cellcolor[HTML]{D8756C}94.95 & 99.33 & 0.1292 & \cellcolor[HTML]{D8756C}0.1460 & \cellcolor[HTML]{D8756C}0.7903  & \cellcolor[HTML]{D8756C}83.90 & 95.87 & 98.89\\ 
Classroom & \cellcolor[HTML]{A2D86C}0.0719 & 0.0282 & 0.2873& \cellcolor[HTML]{A2D86C}95.37 & 99.43 & \cellcolor[HTML]{A2D86C}99.91 & 0.0744 & 0.0344 & 0.3034 & 94.73 & \cellcolor[HTML]{A2D86C}98.94 & 99.72\\
Library  & 0.1163 & 0.0875 & 0.6274 & 87.34 & 96.93 & 98.99  & 0.1258 & 0.1311 & 0.6885  & 85.22 & \cellcolor[HTML]{D8756C}95.34 & 98.62\\
Kitchen & 0.1256 & 0.0589 & 0.2825  & 87.10 & 96.70 & \cellcolor[HTML]{D8756C}98.00 & 0.0958 & 0.0538 & 0.2649 & 92.56 & 96.36 & 98.64\\ 
Playroom  & 0.0790 & 0.0300 & 0.2508 & 94.21 & 98.09 & 99.88 & \cellcolor[HTML]{A2D86C}0.0735 & \cellcolor[HTML]{A2D86C}0.0313 & 0.2574 & \cellcolor[HTML]{A2D86C}95.17 & 98.38 & 99.83\\ 
Living room & 0.0862 & 0.0390 & 0.3036  & 91.72 & 98.31 & 99.58  & 0.0910 & 0.0510 & 0.3490 & 91.45 & 97.59 & 99.25 \\
Bathroom  & 0.1065 & 0.0358 & \cellcolor[HTML]{A2D86C}0.1880 & 93.64 & 97.64 & 98.17 & 0.0955 & 0.0472 & \cellcolor[HTML]{A2D86C}0.2279 & 92.95 & 97.04 & \cellcolor[HTML]{D8756C}97.98\\
\hlineB{2}
\textbf{Standard Deviation} & \underline{0.0250} & \underline{0.0465} & \underline{0.2448} & \underline{6.57} & \underline{1.83} & \textbf{0.85} & \textbf{0.0211} & \textbf{0.0444} & \textbf{0.1947} & \textbf{4.83} & \textbf{1.57} & \underline{1.00} \\
Value Max & 0.1565 & 0.1701 & 0.9157 & 95.37 & 99.44 & 99.91 & 0.1313 & 0.1460 & 0.7903 & 95.17 & 98.94 & 99.83\\
Value Min & 0.0719 & 0.0253 & 0.1880 & 77.19 & 94.95 & 98.00 & 0.0735 & 0.0313 & 0.2279 & 83.06 & 95.34 & 97.98\\
\hline
\end{tabular}

\begin{tabular}{|l|l|l|}
\hline
\cellcolor[HTML]{FCB381} Easy/ Hard Summary & \multicolumn{1}{c|}{\cellcolor[HTML]{99CCFF} ZoeDepth(N)} & \multicolumn{1}{c|}{\cellcolor[HTML]{99CCFF} VPD} \\
\hline
\multicolumn{1}{|c|} {\cellcolor[HTML]{FAE5D3} \textbf{Easy type}} & \cellcolor[HTML]{FAE5D3}bathroom, private room, playroom, classroom& \cellcolor[HTML]{FAE5D3} bathroom, private room, kitchen, classroom\\
\hline
\multicolumn{1}{|c|} {\cellcolor[HTML]{D5F5E3} \textbf{Hard type}} & \cellcolor[HTML]{D5F5E3} large room, lounge, library, hallway &\cellcolor[HTML]{D5F5E3} large room, lounge, library, hallway \\
\hline
\end{tabular}

\vspace{2pt}

\begin{tabular}{|c|c|c|c|c|c|c?c|c|c|c|c|c|}
\hline
\cellcolor[HTML]{99CCFF} $\dagger$ \textbf{M\&LS-Pre} & \multicolumn{6}{c?}{\cellcolor[HTML]{99CCFF} DepthAnything Results} & \multicolumn{6}{c|}{\cellcolor[HTML]{99CCFF}Unidepth Results}\\

\cellcolor[HTML]{FCB381} Type Breakdown & \cellcolor[HTML]{FAE5D3} AbsRel & \cellcolor[HTML]{FAE5D3} SqRel & \cellcolor[HTML]{FAE5D3} RMSE & \cellcolor[HTML]{D5F5E3} $\delta_1$ & \cellcolor[HTML]{D5F5E3} $\delta_2$ & \cellcolor[HTML]{D5F5E3} $\delta_3$ &\cellcolor[HTML]{FAE5D3} AbsRel & \cellcolor[HTML]{FAE5D3} SqRel & \cellcolor[HTML]{FAE5D3} RMSE & \cellcolor[HTML]{D5F5E3} $\delta_1$ & \cellcolor[HTML]{D5F5E3} $\delta_2$ & \cellcolor[HTML]{D5F5E3} $\delta_3$\\
\hline
Private room & \cellcolor[HTML]{A2D86C}0.0836  & \cellcolor[HTML]{A2D86C}0.0297 & 0.2500 & 91.87  & 98.78 & \cellcolor[HTML]{A2D86C}99.80  & 0.0609 & \cellcolor[HTML]{A2D86C}0.0163 & 0.1805 & \cellcolor[HTML]{A2D86C}96.94 & 99.50 & 99.89 \\
Office  & 0.0990  & 0.0454 & 0.3130 & 88.12  & 97.61 & 99.45  & 0.0788 & 0.0322 & 0.2608  & 93.25 & 98.44 & 99.64\\ 
Hallway&  \cellcolor[HTML]{D8756C}0.1074  & 0.0683 & 0.4726 & \cellcolor[HTML]{D8756C}87.02  & \cellcolor[HTML]{D8756C}96.55 & 99.05 & 0.0871 & 0.0566 & 0.4608  & \cellcolor[HTML]{D8756C}90.83 & \cellcolor[HTML]{D8756C}97.14 & 99.10\\ 
Lounge& 0.1053  & 0.0920 & 0.6310 & 88.34  & 97.69 & 99.33 & 0.0860 & 0.0663 & 0.5312  & 93.24 & 98.39 & 99.41\\
Meeting room  &  0.0899  & 0.0462 & 0.3591 & 90.28  & 98.52 & 99.76   & 0.0764 & 0.0330 & 0.2992  & 94.66 & 99.12 & 99.79\\ 
Large room  & 0.1051  & \cellcolor[HTML]{D8756C}0.0974 & \cellcolor[HTML]{D8756C}0.6860 & 88.61  & 97.91 & 99.27 & \cellcolor[HTML]{D8756C}0.0871 & \cellcolor[HTML]{D8756C}0.0688 & \cellcolor[HTML]{D8756C}0.5370  & 94.10 & 98.46 & 99.26\\ 
Classroom & 0.0870  & 0.0426 & 0.3465 & 90.24  & 98.61 & 99.81 & 0.0601 & 0.0233 & 0.2589  & 96.67 & \cellcolor[HTML]{A2D86C}99.52 & \cellcolor[HTML]{A2D86C}99.93\\
Library  & 0.0947  & 0.0641 & 0.5150 & 90.08  & 98.09 & 99.43  & 0.0800 & 0.0498 & 0.4505  & 93.63 & 98.37 & 99.40\\
Kitchen &  0.0863  & 0.0390 & 0.2480 & 91.97  & 97.99 & 99.13 & 0.0662 & 0.0235 & 0.2027  & 95.42 & 98.80 & 99.34\\ 
Playroom  & 0.0871  & 0.0416 & 0.3453 & 91.23  & 98.38 & 99.62 & \cellcolor[HTML]{A2D86C}0.0544 & 0.0182 & 0.1983  & 95.60 & 99.30 & 99.89\\ 
Living room & 0.0921  & 0.0472 & 0.3448 & 90.26  & 97.80 & 99.47 & 0.0670 & 0.0312 & 0.2758  & 94.87 & 98.54 & 99.58 \\
Bathroom  & 0.0860  & 0.0337 & \cellcolor[HTML]{A2D86C}0.1965 & \cellcolor[HTML]{A2D86C}92.71  & \cellcolor[HTML]{A2D86C}99.72 & \cellcolor[HTML]{D8756C}98.76 & 0.0765 & 0.0252 & \cellcolor[HTML]{A2D86C}0.1600  & 96.40 & 98.00 & \cellcolor[HTML]{D8756C}98.52\\
\hlineB{2}
\textbf{Standard Deviation} & \textbf{0.0085} & \underline{0.0257} & \textbf{0.1470} & \textbf{1.56} & \underline{0.87} & \textbf{0.36} & \underline{0.0118} & \textbf{0.0204} & \underline{0.1471} & \underline{1.96} & \textbf{0.72} & \underline{0.58} \\
Value Max & 0.1074 & 0.0974 & 0.6860 & 92.71 & 99.72 & 99.80 & 0.0871 & 0.0688 & 0.5370 & 96.94 & 99.52 & 99.93\\
Value Min & 0.0836 & 0.0297 & 0.1965 & 87.02 & 96.55 & 98.76 & 0.0544 & 0.0163 & 0.1600 & 90.83 & 97.14 & 98.52\\
\hline
\end{tabular}

\begin{tabular}{|l|l|l|l|}
\hline
\cellcolor[HTML]{FCB381} Easy/ Hard Summary & \multicolumn{1}{c|}{\cellcolor[HTML]{99CCFF} DepthAnything} & \multicolumn{1}{c|}{\cellcolor[HTML]{99CCFF} Unidepth} \\
\hline
\multicolumn{1}{|c|} {\cellcolor[HTML]{FAE5D3} \textbf{Easy type}} & \cellcolor[HTML]{FAE5D3}bathroom, private room, kitchen& \cellcolor[HTML]{FAE5D3} bathroom, private room, kitchen, classroom, playroom\\
\hline
\multicolumn{1}{|c|} {\cellcolor[HTML]{D5F5E3} \textbf{Hard type}} & \cellcolor[HTML]{D5F5E3} large room, lounge, library, meeting room &\cellcolor[HTML]{D5F5E3} hallway, lounge, library, office \\
\hline
\end{tabular}
\vspace{-13pt}
\end{table*}

\textbf{Analysis}. Table~\ref{table:2} shows the comparison for overall performance. VPD, MIM, ZoeDepth, DepthAnything, and Unidepth are among the top-performing methods.
They commonly adopt transformers ~\cite{dosovitskiy2020image,bao2021beit,liu2021Swin} and learn better representations by large models.
VPD learns from Stable Diffusion~\cite{rombach2022high} and utilizes its pretrained image priors to help learn geometry. 
ZoeDepth and DepthAnything are first pretrained on a mixture of datasets, including indoor and outdoor scenes, and then fine-tuned on NYUv2 for the benchmark.
In contrast, Unidepth does not require fine-tuning, and its paper reports zero-shot performance on NYUv2 that even slightly outperforms fine-tuned DepthAnything.
One can observe that Unidepth has larger performance gaps to others on the benchmark where all the methods are evaluated by zero-shot performance.

\begin{figure*}[bt!]
    \centering
    \includegraphics[width=0.99\linewidth]{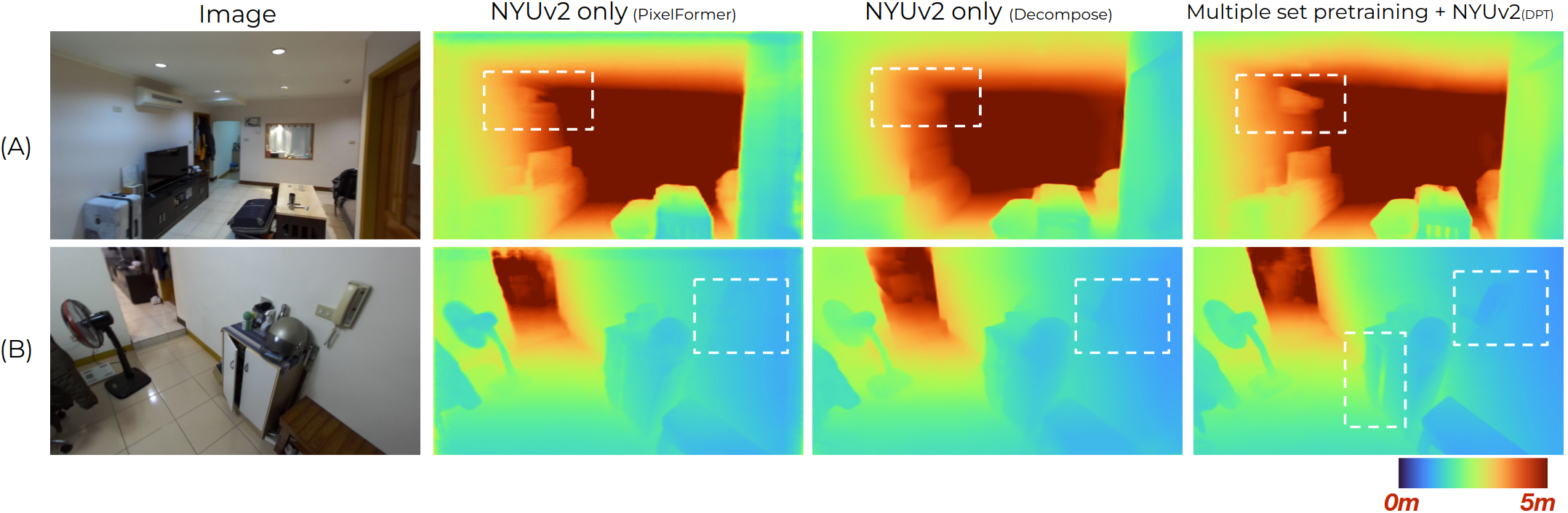}
    \vspace{-7pt}
    \caption{\textbf{Zero-shot comparison between training strategies.}    
    The rooms are in oriental styles where the wall-hanging phone and air-conditioner are mostly exclusive to the cultural style. 
    Compared with training on NYUv2 only, the multiple dataset pretraining shows higher robustness to unseen spaces, even though DPT is inferior to PixelFormer and Decompose on the standard NYUv2 dataset. Zero-shot evaluation on InSpaceType helps figure out advantages that cannot be shown if focusing on the single benchmark.}
    \vspace{-7pt}
    \label{fig_diff}
\end{figure*}

\textbf{Discussion on robustness}. Even an early method, DPT, which is also pretrained on multiple datasets and fine-tuned on NYUv2, can slightly outperform several later methods in terms of AbsRel, such as PixelFormer, IronDepth, or Decomposition, which actually won over DPT on the NYUv2 benchmark. 
Fig.~\ref{fig_diff} shows a visual comparison using oriental-style rooms. The type did not appear in NYUv2 or DPT's training sets.
Compared with single-dataset training such as PixelFormer and Decompose, DPT generalizes better to unseen spaces. 
The advantage in robustness cannot be shown if fixating on a single benchmark.


\textbf{[\RNum{2}: Breakdowns by space types]} Compared with overall performance that follows natural distribution, the aim here is to leave out the effect of number of samples in a type and break down performance by types to discover performance variances.
Our tool shows error and accuracy metrics for each node in our hierarchical graph. See examples in the supplementary. Level-0: overall performance; Level-1: household, workspace, campus, and functional space; Level-2: private room, office, hallway, lounge, meeting room, large room, classroom, library, kitchen, playroom, living room, and bathroom; Level-3 contains more detailed types from level-2 if dividable. There are 26 leaf nodes in total.
We focus on the 12 types in level-2 in the main paper and include the full type reports in the supplementary for reference.
\textit{N-only} contains the top methods, MIM and PixelFormer, among those trained on RGBD data only from NYUv2; \textit{M\&LS-Pre} contains the top methods, VPD, ZoeDepth, DepthAnything, and Unidepth, among those trained on multiple depth datasets or learned from large-scale pretraining.

\textbf{Performance variance analysis}. Table~\ref{table:25} shows performance on each space type.
For N-only, though MIM is among the leading methods in Table~\ref{table:2} overall results and outperforms PixelFormer, its standard deviation and difference between max and min are much larger than PixelFormer, indicating a high imbalance. 
The gap between the best and worst (RMSE and $\delta_1$) is prominent, which goes by (0.2466, 94.54) and (1.3915, 54.93).
For M\&LS-Pre in the second block, similarly, ZoeDepth has better overall performance than VPD in Table~\ref{table:2} but has larger performance variances shown by the standard deviation. 
Likewise, comparing DepthAnything and Unidepth, though DepthAnyting has lower overall performance, its performance variances are slightly lower than Unidepth in terms of standard deviation for RMSE and $\delta_1$, showing lower imbalance.

\textbf{Easy/Hard types}. We also enumerate easy and hard types for each method based on the concordance in top-5 lowest error/ highest accuracy (easy) and the opposite (hard). 
The easy and hard types are mostly consistent, showing the natural space type distribution underlying training data. 
The trained models then follow the distribution and favor the head types, such as private room and bathroom, and against the tailed types, such as library and large room. 
All the methods more or less have clear easy and hard types indicating bias, and some even have obvious gaps between head and tailed types from the numerical results.
The lists used as references reveal where the models excel and provide guidance on their proper usage. One may expect degraded performance on the tailed types and need to curate a dataset to fine-tune to make up the gaps.
Note that \textbf{we do not claim clear easy and hard types are always detrimental}, as some applications may require focusing on specific types.
\textit{We argue one needs to attend to a pretrained model's underlying bias, understand its advantages and weakness, and be careful about overall performance reported on a single benchmark where robustness may not be fully validated.}

\textbf{[\RNum{3}: More training sets and discussion on synthetic data curation]} Going beyond the NYUv2 benchmark, there are several recent larger monocular indoor depth datasets for training purposes. 
We investigate the datasets by using models pretrained on them to zero-shot evaluate on InSpaceType. We study SimSIN and UniSIN \cite{wu2022toward}; the former is synthetic, and the latter is real. They are both introduced by DistDepth \cite{wu2022toward} with a focus on self-supervised learning. We adopt DistDepth's released models for evaluation. We also study Hypersim~\cite{roberts2021hypersim}, which is purely synthetic and rendered from crafted 3D CAD indoor environments, and we use its pretrained ConvNext-Base model~\cite{liu2022convnet}. Table~\ref{table:3} shows the performance and type breakdowns. Due to the synthetic-to-real and self-supervised learning challenges, their overall performance is lower than training on NYUv2 in Table~\ref{benchmark}.

\textbf{Dataset characteristics and analysis}. SimSIN: The dataset was built by rendering from an agent-navigating simulator~\cite{savva2019habitat} with scanned but incomplete household meshes from the datasets~\cite{straub2019replica,chang2017matterport3d,ramakrishnan2021habitat}. Thus, it may favor household spaces and bias against other types, such as workspace or campus scenes;
UniSIN: It contains diverse real scenes covering workspace, household, and campus spaces and thus shows fewer hard types;
Hypersim: It renders diverse indoor 3D CAD spaces with fine textures.
However, many rendered scenes are highly under-/over-exposed, and many frames focus on small objects or ceilings that cut off cues for depth hints. 
Hypersim also contains household spaces as the majority and further includes several open or unbounded spaces.

Both SimSIN and Hypersim are synthetic, and their data curation focuses more on head types, especially private room and living room as the most common application contexts. Training on the datasets may underrepresent certain tailed types such as library, lounge, or hallway. 
It is worth to mention hallway is a common but easily overlooked type in synthetic datasets, especially for those who render every 3D CAD space separately without considering space connection, such as Hypersim, which shows low $\delta_1$ compared to real scanned UniSIN.



\begin{figure}[bt!]
    \centering
    \includegraphics[width=1.00\linewidth]{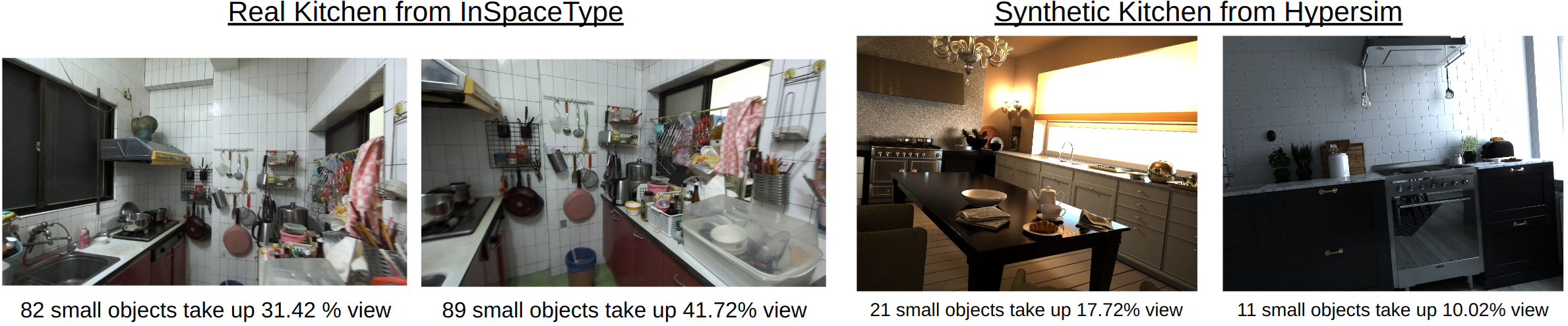}
    \vspace{-14pt}
    \caption{Real kitchen scenes contain many small objects in large areas. However, the synthetic data cannot match the complexity hindering synthetic-to-real performance.}
    \vspace{-3pt}
    \label{fig_obj_validation}
\end{figure}

\textbf{Scene complexity for synthetic v.s. real data}.
One can find kitchen is a special type in SimSIN and Hypersim, which is of lower RMSE but also very low accuracy $\delta_1$/ high AbsRel compared with other types. 
Kitchens are small-range with naturally lower RMSE if a network predicts near ranges. 
However, the lower $\delta_1$ and higher AbsRel indicate detailed surface depth for objects is not accurately predicted. 
This is because kitchen scenes often abound with cluttered and small objects spread across the view (Fig.~\ref{fig_obj_validation}). Strategies of synthetic data curation, such as rendering from scanned but incomplete meshes (SimSIN\footnote{small or thin objects are usually skewed in shape or incomplete due to limited laser scan resolutions.}) or from crafted 3D CAD indoor environments (Hypersim\footnote{most 3D CAD models favor neat and organized spaces, and crafting cluttered/ small objects takes manual efforts.}), cannot faithfully reflect the high complexity of cluttered and small objects in real scenes. 
The study highlights common deficiencies in synthetic data curation for synthetic-to-real transfer. One needs to consider scene complexity while creating synthetic data to achieve better in real scenes.

\begin{table}[bt!]
\centering
\vspace{0pt}
\caption{\textbf{Space type breakdowns and characteristics} for more real and synthetic datasets.} 
\vspace{3pt}
\scriptsize

\label{table:3}

\begin{tabular}{|c|c|c|c|c|c|c|c|c|c|}
\hline
\cellcolor[HTML]{DDCCFA}Dataset Study& \multicolumn{3}{c|}{\cellcolor[HTML]{DDCCFA} SimSIN}  & \multicolumn{3}{c|}{\cellcolor[HTML]{DDCCFA} UniSIN}   & \multicolumn{3}{c|}{\cellcolor[HTML]{DDCCFA} Hypersim}\\

\cellcolor[HTML]{FCB381} Type Breakdown & \cellcolor[HTML]{FAE5D3} AbsRel & \cellcolor[HTML]{FAE5D3} RMSE & \cellcolor[HTML]{D5F5E3} $\delta_1$ & \cellcolor[HTML]{FAE5D3} AbsRel & \cellcolor[HTML]{FAE5D3} RMSE & \cellcolor[HTML]{D5F5E3} $\delta_1$ & \cellcolor[HTML]{FAE5D3} AbsRel & \cellcolor[HTML]{FAE5D3} RMSE & \cellcolor[HTML]{D5F5E3} $\delta_1$\\
\hline
Private room  & \cellcolor[HTML]{A2D86C}0.1509  & 0.4447 & \cellcolor[HTML]{A2D86C}79.36 & 0.1736  & 0.4527 & \cellcolor[HTML]{D8756C}72.89 & 0.1321 & 0.3376 & 84.55  \\
Office & 0.1812   & 0.5789 & 74.01 & 0.1650  & 0.5016 & 77.12  & 0.1678  & 0.4916 & 76.01\\ 
Hallway & 0.1597  & 0.6324 & 78.12 & 0.1258 & 0.4970 & 85.13  & 0.2126  & 0.8347 & \textbf{65.67} \\ 
Lounge & 0.1841  & 0.9037 & 73.86 & 0.1375  & 0.7347 & 83.08 & 0.1937  & 0.9403 & 71.31\\ 
Meeting room & 0.1962 & 0.9417  & 66.91 & 0.1295 & 0.5938 & 84.17& 0.1315  & 0.5333  & 81.49\\ 
Large room  & 0.1842   & \cellcolor[HTML]{D8756C}1.0727 & 72.79 & 0.1607& \cellcolor[HTML]{D8756C}0.9449 & 75.36 & \cellcolor[HTML]{D8756C}0.2525  & \cellcolor[HTML]{D8756C}1.3897  & \cellcolor[HTML]{D8756C}56.48 \\ 
Classroom  & 0.2069 & 1.0292  & 67.11 & 0.1301 & 0.6072 & 85.01  & \cellcolor[HTML]{A2D86C}0.1258  & 0.4700  & \cellcolor[HTML]{A2D86C}85.07  \\
Library & 0.1857   & 0.8307 & 75.14 & \cellcolor[HTML]{A2D86C}0.1218  & 0.6921 & \cellcolor[HTML]{A2D86C}85.88& 0.1766  & 0.8654  & 73.53  \\ 
Kitchen  & \cellcolor[HTML]{D8756C}0.2524   & 0.5649 & \cellcolor[HTML]{D8756C}59.22 & \cellcolor[HTML]{D8756C}0.2122  & 0.4950 & 75.66 & 0.2417  & 0.4347 & \textbf{62.81}\\ 
Playroom & 0.1597   & 0.5946 & 75.59 & 0.1475 & 0.4710 & 78.99& 0.1629  & 0.5663  & 77.09  \\ 
Living room & 0.1600   & 0.5284 & 77.05  & 0.1642 & 0.5108 & 76.51 & 0.1485  & 0.4519 & 80.91\\
Bathroom & 0.1751   & \cellcolor[HTML]{A2D86C}0.3900 & 74.22& 0.1381  & \cellcolor[HTML]{A2D86C}0.2163 & 84.89 & 0.1648  & \cellcolor[HTML]{A2D86C}0.2908 & 76.11 \\
\hline
All & 0.1746 & 0.6877 & 74.72& 0.1426 & 0.5414 & 80.23 & 0.1592  & 0.5803 & 77.95\\
\hline
\end{tabular}

\begin{tabular}{|l|l|l|l|}

\hline
\cellcolor[HTML]{FCB381} Summary & \multicolumn{1}{c|}{\cellcolor[HTML]{DDCCFA} SimSIN} & \multicolumn{1}{c|}{\cellcolor[HTML]{DDCCFA} UniSIN} & \multicolumn{1}{c|}{\cellcolor[HTML]{DDCCFA} Hypersim} \\
\hline

\multicolumn{1}{|c|} {\cellcolor[HTML]{FAE5D3} \textbf{Easy type}} & \scriptsize \cellcolor[HTML]{FAE5D3} living room, private room  & \cellcolor[HTML]{FAE5D3} bathroom, hallway  & \cellcolor[HTML]{FAE5D3} private room, living room, classroom \\
\hline

\multicolumn{1}{|c|} {\cellcolor[HTML]{D5F5E3} \textbf{Hard type}} & \cellcolor[HTML]{D5F5E3} large room, classroom, meeting room, lounge & \cellcolor[HTML]{D5F5E3} large room& \cellcolor[HTML]{D5F5E3} large room, library, hallway, lounge  \\
\hline
\end{tabular}

\vspace{-3pt}
\end{table}

\textbf{[\RNum{4}: Dataset ablation for generalization study]}  
Aside from the above cross-dataset generalization experiments, 
consider a case when a space type is missed in a dataset, how hard is it, and what are key factors to generalize to unseen types? 
The study uses all the collected InSpaceType data to form a split with about 40K RGBD pairs, where the previous 1260 evaluation pairs and their nearby 2 frames are excluded.
This 40K split is divided into 3 groups by similarity in functions.  
G1 (household space): private room, kitchen, living room, and bathroom; 
G2 (workspace and campus): office, hallway, meeting room, and classroom; 
G3 (large functional space): lounge, large room, playroom, and library.
Then we train models 
\footnote{without loss of generality, we use general-purpose ConvNeXt-L and standard $L_2$, smoothness, and SILog loss. We train by 30 epochs, which is adequate for convergence regarding loss. The learning rate is set at 0.0001, momentum at (0.9, 0.999), and weight decay at 0.01 with an AdamW optimizer.}
on each group and evaluate on the 1260-pair evaluation set. For instance, G1$\to$G2 means training on G1, where a model only has seen G1's type, 
and testing on data of G2's types in the evaluation set. Table~\ref{table:15} provides the results.


\begin{table}[bt!]
\scriptsize
\centering

\caption{\textbf{Cross-group generalization.} The header $\to$ specifies a source training group, and the rows below show testing groups and results. 
}
\vspace{2pt}

\begin{tabular}{|c|c|c|c|c|c|c|c|c|c|}
\hline
& \multicolumn{3}{c}{\cellcolor[HTML]{FFFE65} G1$\to$}  & \multicolumn{3}{c}{\cellcolor[HTML]{F7BCDC} G2$\to$}   & \multicolumn{3}{c}{\cellcolor[HTML]{EBDEF0} G3$\to$}\\
\cellcolor[HTML]{99CCFF} Test group & \cellcolor[HTML]{FAE5D3} AbsRel & \cellcolor[HTML]{FAE5D3} RMSE & \cellcolor[HTML]{D5F5E3} $\delta_1$ & \cellcolor[HTML]{FAE5D3} AbsRel & \cellcolor[HTML]{FAE5D3} RMSE & \cellcolor[HTML]{D5F5E3} $\delta_1$ & \cellcolor[HTML]{FAE5D3} AbsRel & \cellcolor[HTML]{FAE5D3} RMSE & \cellcolor[HTML]{D5F5E3} $\delta_1$  \\
\hline
G1 & 0.0735 & 0.1851 & 95.42 & 0.1415  & 0.3478 & 82.44 & 0.1886 & 0.4744 & 73.01 \\
G2 & 0.1588 & 0.5488 & 79.10 & 0.0790  & 0.2470 & 94.79 & 0.1945 & 0.6101 & 72.88\\
G3 & 0.2644 & 1.1552 & 62.05 & 0.2107 & 0.9400 & 70.18  & 0.0729 & 0.3470 & 95.65\\
\hline
\end{tabular}

\label{table:15}
\vspace{-7pt}
\end{table}

Generalization to unseen types (e.g., G1$\to$G3) is more challenging than seen types (e.g., G1$\to$G1). One can observe that household space (G1) and workspace (G2) are easier to generalize to each other than generalize to large functional space (G3), and vice versa for G3$\to$G1/G2.
G1 and G2 are mainly distinct in object arrangements and appearance, whereas G1 and G3 mainly differ in depth ranges. The average maximal depth is 3.78m for G1, 5.49m for G2, and 12.08m for G3.
The results indicate the scene scale is the key factor. Training on one group inherently introduces a bias towards the output depth ranges. When presented with a scene in a much different range, the model struggles to accurately estimate the corresponding depth range. 
\vspace{-7pt}

\vspace{-4pt}
\section{Conclusion}
\label{conclusion}
\vspace{-2pt}
The work pioneers studying space types in indoor monocular depth for practical purposes, especially with the advent of many large models, but the evaluation and quality assessment still primarily focus on a single and older benchmark. 
First, we present novel InSpaceType Dataset that meets the high-resolution and high-quality RGBD data requirements for cutting-edge applications in AR/VR displays and indoor robotics.
Previous works focusing on methods may overlook performance variances. We use InSpaceType to study 13 recent high-performing methods and analyze their zero-shot cross-dataset performance for both overall results and performance variances across space types. Even some top methods have severe imbalance, and some methods are actually less imbalanced than higher-performing ones.   

We extend our analysis to more synthetic and real datasets, including SimSIN, UniSIN, and Hypersim, to reveal their bias and guide proper usage.
Especially, current synthetic data curation may not faithfully reflect the real-world high complexity in cluttered and small objects, and we suggest the best practice. Further, they may miss some common types like hallway if rendering single 3D CAD spaces separately. We further do ablation on InSpaceType and find space scale is the key factor that hinders generalization.
As part of our contribution, our released tools for both research and practical aids, including codes and datasets, can diagnose a pretrained model and show its hierarchical performance breakdown. Overall, this work underscores the importance of considering performance variances in the practical deployment of models, a crucial aspect often overlooked in the field.

{\small
\bibliography{egbib}

\begin{thebibliography}{54}
\providecommand{\natexlab}[1]{#1}
\providecommand{\url}[1]{\texttt{#1}}
\expandafter\ifx\csname urlstyle\endcsname\relax
  \providecommand{\doi}[1]{doi: #1}\else
  \providecommand{\doi}{doi: \begingroup \urlstyle{rm}\Url}\fi

\bibitem[Kin()]{Kinect}
Hardware.cast technical publications series.
\newblock \url{https://gmv.cast.uark.edu/scanning/hardware/microsoft-kinect-resourceshardware}.

\bibitem[Agarwal and Arora(2023)]{agarwal2023attention}
Ashutosh Agarwal and Chetan Arora.
\newblock Attention attention everywhere: Monocular depth prediction with skip attention.
\newblock In \emph{WACV}, 2023.

\bibitem[Bae et~al.(2022)Bae, Budvytis, and Cipolla]{bae2022irondepth}
Gwangbin Bae, Ignas Budvytis, and Roberto Cipolla.
\newblock Irondepth: Iterative refinement of single-view depth using surface normal and its uncertainty.
\newblock \emph{BMVC}, 2022.

\bibitem[Bao et~al.(2021)Bao, Dong, Piao, and Wei]{bao2021beit}
Hangbo Bao, Li~Dong, Songhao Piao, and Furu Wei.
\newblock Beit: Bert pre-training of image transformers.
\newblock \emph{ICLR}, 2021.

\bibitem[Bhat et~al.(2021)Bhat, Alhashim, and Wonka]{bhat2021adabins}
Shariq~Farooq Bhat, Ibraheem Alhashim, and Peter Wonka.
\newblock Adabins: Depth estimation using adaptive bins.
\newblock In \emph{CVPR}, 2021.

\bibitem[Bhat et~al.(2022)Bhat, Alhashim, and Wonka]{bhat2022localbins}
Shariq~Farooq Bhat, Ibraheem Alhashim, and Peter Wonka.
\newblock Localbins: Improving depth estimation by learning local distributions.
\newblock In \emph{ECCV}, 2022.

\bibitem[Bhat et~al.(2023)Bhat, Birkl, Wofk, Wonka, and M{\"u}ller]{bhat2023zoedepth}
Shariq~Farooq Bhat, Reiner Birkl, Diana Wofk, Peter Wonka, and Matthias M{\"u}ller.
\newblock Zoedepth: Zero-shot transfer by combining relative and metric depth.
\newblock \emph{arXiv preprint arXiv:2302.12288}, 2023.

\bibitem[Bian et~al.(2021)Bian, Zhan, Wang, Chin, Shen, and Reid]{bian2021auto}
Jia-Wang Bian, Huangying Zhan, Naiyan Wang, Tat-Jun Chin, Chunhua Shen, and Ian Reid.
\newblock Auto-rectify network for unsupervised indoor depth estimation.
\newblock \emph{TPAMI}, 2021.

\bibitem[Chang et~al.(2017)Chang, Dai, Funkhouser, Halber, Niebner, Savva, Song, Zeng, and Zhang]{chang2017matterport3d}
Angel Chang, Angela Dai, Thomas Funkhouser, Maciej Halber, Matthias Niebner, Manolis Savva, Shuran Song, Andy Zeng, and Yinda Zhang.
\newblock {Matterport3D}: Learning from {RGB-D} data in indoor environments.
\newblock In \emph{3DV}, 2017.

\bibitem[Deng et~al.(2022)Deng, Liu, Zhu, and Ramanan]{deng2022depth}
Kangle Deng, Andrew Liu, Jun-Yan Zhu, and Deva Ramanan.
\newblock Depth-supervised nerf: Fewer views and faster training for free.
\newblock In \emph{CVPR}, 2022.

\bibitem[Diaz et~al.(2017)Diaz, Walker, Szafir, and Szafir]{diaz2017designing}
Catherine Diaz, Michael Walker, Danielle~Albers Szafir, and Daniel Szafir.
\newblock Designing for depth perceptions in augmented reality.
\newblock In \emph{2017 IEEE international symposium on mixed and augmented reality (ISMAR)}, 2017.

\bibitem[Dosovitskiy et~al.(2020)Dosovitskiy, Beyer, Kolesnikov, Weissenborn, Zhai, Unterthiner, Dehghani, Minderer, Heigold, Gelly, et~al.]{dosovitskiy2020image}
Alexey Dosovitskiy, Lucas Beyer, Alexander Kolesnikov, Dirk Weissenborn, Xiaohua Zhai, Thomas Unterthiner, Mostafa Dehghani, Matthias Minderer, Georg Heigold, Sylvain Gelly, et~al.
\newblock An image is worth 16x16 words: Transformers for image recognition at scale.
\newblock In \emph{ICLR}, 2020.

\bibitem[Flacco et~al.(2012)Flacco, Kr{\"o}ger, De~Luca, and Khatib]{flacco2012depth}
Fabrizio Flacco, Torsten Kr{\"o}ger, Alessandro De~Luca, and Oussama Khatib.
\newblock A depth space approach to human-robot collision avoidance.
\newblock In \emph{ICRA}, 2012.

\bibitem[Fu et~al.(2018)Fu, Gong, Wang, Batmanghelich, and Tao]{fu2018deep}
Huan Fu, Mingming Gong, Chaohui Wang, Kayhan Batmanghelich, and Dacheng Tao.
\newblock Deep ordinal regression network for monocular depth estimation.
\newblock In \emph{CVPR}, 2018.

\bibitem[Jiang et~al.(2021)Jiang, Ding, Hu, and Huang]{jiang2021plnet}
Hualie Jiang, Laiyan Ding, Junjie Hu, and Rui Huang.
\newblock Plnet: Plane and line priors for unsupervised indoor depth estimation.
\newblock In \emph{3DV}, 2021.

\bibitem[Jun et~al.(2022)Jun, Lee, Lee, and Kim]{jun2022depth}
Jinyoung Jun, Jae-Han Lee, Chul Lee, and Chang-Su Kim.
\newblock Depth map decomposition for monocular depth estimation.
\newblock In \emph{ECCV}, 2022.

\bibitem[Kim et~al.(2022)Kim, Ga, Ahn, Joo, Chun, and Kim]{kim2022global}
Doyeon Kim, Woonghyun Ga, Pyungwhan Ahn, Donggyu Joo, Sehwan Chun, and Junmo Kim.
\newblock Global-local path networks for monocular depth estimation with vertical cutdepth.
\newblock \emph{arXiv:2201.07436}, 2022.

\bibitem[Koch et~al.(2018)Koch, Liebel, Fraundorfer, and Korner]{koch2018evaluation}
Tobias Koch, Lukas Liebel, Friedrich Fraundorfer, and Marco Korner.
\newblock Evaluation of cnn-based single-image depth estimation methods.
\newblock In \emph{ECCVW}, 2018.

\bibitem[Lee et~al.(2019)Lee, Han, Ko, and Suh]{lee2019big}
Jin~Han Lee, Myung-Kyu Han, Dong~Wook Ko, and Il~Hong Suh.
\newblock From big to small: Multi-scale local planar guidance for monocular depth estimation.
\newblock \emph{arXiv preprint arXiv:1907.10326}, 2019.

\bibitem[Lee and Ahn(2020)]{lee2020real}
Junwoo Lee and Bummo Ahn.
\newblock Real-time human action recognition with a low-cost rgb camera and mobile robot platform.
\newblock \emph{Sensors}, 2020.

\bibitem[Li et~al.(2021)Li, Huang, Liu, Zou, and Yu]{li2021structdepth}
Boying Li, Yuan Huang, Zeyu Liu, Danping Zou, and Wenxian Yu.
\newblock Structdepth: Leveraging the structural regularities for self-supervised indoor depth estimation.
\newblock In \emph{ICCV}, 2021.

\bibitem[Li et~al.(2022{\natexlab{a}})Li, Wang, Liu, and Zhao]{li2022real}
Wang Li, Junfeng Wang, Maoding Liu, and Shiwen Zhao.
\newblock Real-time occlusion handling for augmented reality assistance assembly systems with monocular images.
\newblock \emph{Journal of Manufacturing Systems}, 2022{\natexlab{a}}.

\bibitem[Li et~al.(2022{\natexlab{b}})Li, Chen, Liu, and Jiang]{li2022depthformer}
Zhenyu Li, Zehui Chen, Xianming Liu, and Junjun Jiang.
\newblock Depthformer: Exploiting long-range correlation and local information for accurate monocular depth estimation.
\newblock \emph{arXiv preprint arXiv:2203.14211}, 2022{\natexlab{b}}.

\bibitem[Li et~al.(2022{\natexlab{c}})Li, Wang, Liu, and Jiang]{li2022binsformer}
Zhenyu Li, Xuyang Wang, Xianming Liu, and Junjun Jiang.
\newblock Binsformer: Revisiting adaptive bins for monocular depth estimation.
\newblock \emph{arXiv preprint arXiv:2204.00987}, 2022{\natexlab{c}}.

\bibitem[Liu et~al.(2023)Liu, Kumar, Gu, Timofte, and Van~Gool]{liuva}
Ce~Liu, Suryansh Kumar, Shuhang Gu, Radu Timofte, and Luc Van~Gool.
\newblock Va-depthnet: A variational approach to single image depth prediction.
\newblock In \emph{ICLR}, 2023.

\bibitem[Liu et~al.(2015)Liu, Shen, and Lin]{liu2015deep}
Fayao Liu, Chunhua Shen, and Guosheng Lin.
\newblock Deep convolutional neural fields for depth estimation from a single image.
\newblock In \emph{CVPR}, 2015.

\bibitem[Liu et~al.(2021)Liu, Lin, Cao, Hu, Wei, Zhang, Lin, and Guo]{liu2021Swin}
Ze~Liu, Yutong Lin, Yue Cao, Han Hu, Yixuan Wei, Zheng Zhang, Stephen Lin, and Baining Guo.
\newblock Swin transformer: Hierarchical vision transformer using shifted windows.
\newblock In \emph{ICCV}, 2021.

\bibitem[Liu et~al.(2022)Liu, Mao, Wu, Feichtenhofer, Darrell, and Xie]{liu2022convnet}
Zhuang Liu, Hanzi Mao, Chao-Yuan Wu, Christoph Feichtenhofer, Trevor Darrell, and Saining Xie.
\newblock A convnet for the 2020s.
\newblock \emph{CVPR}, 2022.

\bibitem[Luo et~al.(2020)Luo, Huang, Szeliski, Matzen, and Kopf]{luo2020consistent}
Xuan Luo, Jia-Bin Huang, Richard Szeliski, Kevin Matzen, and Johannes Kopf.
\newblock Consistent video depth estimation.
\newblock \emph{ACM Transactions on Graphics (TOG)}, 2020.

\bibitem[Mehringer et~al.(2022)Mehringer, Wirth, Roth, Michelson, and Eskofier]{mehringer2022stereopsis}
Wolfgang Mehringer, Markus Wirth, Daniel Roth, Georg Michelson, and Bjoern~M Eskofier.
\newblock Stereopsis only: Validation of a monocular depth cues reduced gamified virtual reality with reaction time measurement.
\newblock \emph{IEEE Transactions on Visualization and Computer Graphics}, 2022.

\bibitem[Ning et~al.(2023)Ning, Li, Zhang, Wang, Geng, Dai, He, and Hu]{Ning_2023_ICCV}
Jia Ning, Chen Li, Zheng Zhang, Chunyu Wang, Zigang Geng, Qi~Dai, Kun He, and Han Hu.
\newblock All in tokens: Unifying output space of visual tasks via soft token.
\newblock In \emph{ICCV}, 2023.

\bibitem[Piccinelli et~al.(2024)Piccinelli, Yang, Sakaridis, Segu, Li, Van~Gool, and Yu]{piccinelli2024unidepth}
Luigi Piccinelli, Yung-Hsu Yang, Christos Sakaridis, Mattia Segu, Siyuan Li, Luc Van~Gool, and Fisher Yu.
\newblock Unidepth: Universal monocular metric depth estimation.
\newblock In \emph{CVPR}, 2024.

\bibitem[Ramakrishnan et~al.(2021)Ramakrishnan, Gokaslan, Wijmans, Maksymets, Clegg, Turner, Undersander, Galuba, Westbury, Chang, et~al.]{ramakrishnan2021habitat}
Santhosh~K Ramakrishnan, Aaron Gokaslan, Erik Wijmans, Oleksandr Maksymets, Alex Clegg, John Turner, Eric Undersander, Wojciech Galuba, Andrew Westbury, Angel~X Chang, et~al.
\newblock Habitat-matterport {3D} dataset {(HM3D)}: 1000 large-scale {3D} environments for embodied {AI}.
\newblock \emph{NeurIPS Datasets and Benchmarks Track}, 2021.

\bibitem[Ramamonjisoa et~al.(2021)Ramamonjisoa, Firman, Watson, Lepetit, and Turmukhambetov]{ramamonjisoa2021single}
Michael Ramamonjisoa, Michael Firman, Jamie Watson, Vincent Lepetit, and Daniyar Turmukhambetov.
\newblock Single image depth prediction with wavelet decomposition.
\newblock In \emph{CVPR}, 2021.

\bibitem[Ranftl et~al.(2020)Ranftl, Lasinger, Hafner, Schindler, and Koltun]{Ranftl2020}
Ren\'{e} Ranftl, Katrin Lasinger, David Hafner, Konrad Schindler, and Vladlen Koltun.
\newblock Towards robust monocular depth estimation: Mixing datasets for zero-shot cross-dataset transfer.
\newblock \emph{TPAMI}, 2020.

\bibitem[Ranftl et~al.(2021)Ranftl, Bochkovskiy, and Koltun]{Ranftl2021}
Ren\'{e} Ranftl, Alexey Bochkovskiy, and Vladlen Koltun.
\newblock Vision transformers for dense prediction.
\newblock \emph{ICCV}, 2021.

\bibitem[Ricci et~al.(2018)Ricci, Ouyang, Wang, Sebe, et~al.]{ricci2018monocular}
Elisa Ricci, Wanli Ouyang, Xiaogang Wang, Nicu Sebe, et~al.
\newblock Monocular depth estimation using multi-scale continuous crfs as sequential deep networks.
\newblock \emph{TPAMI}, 2018.

\bibitem[Roberts et~al.(2021)Roberts, Ramapuram, Ranjan, Kumar, Bautista, Paczan, Webb, and Susskind]{roberts2021hypersim}
Mike Roberts, Jason Ramapuram, Anurag Ranjan, Atulit Kumar, Miguel~Angel Bautista, Nathan Paczan, Russ Webb, and Joshua~M Susskind.
\newblock Hypersim: A photorealistic synthetic dataset for holistic indoor scene understanding.
\newblock In \emph{ICCV}, 2021.

\bibitem[Rombach et~al.(2022)Rombach, Blattmann, Lorenz, Esser, and Ommer]{rombach2022high}
Robin Rombach, Andreas Blattmann, Dominik Lorenz, Patrick Esser, and Bj{\"o}rn Ommer.
\newblock High-resolution image synthesis with latent diffusion models.
\newblock In \emph{CVPR}, 2022.

\bibitem[Savva et~al.(2019)Savva, Kadian, Maksymets, Zhao, Wijmans, Jain, Straub, Liu, Koltun, Malik, et~al.]{savva2019habitat}
Manolis Savva, Abhishek Kadian, Oleksandr Maksymets, Yili Zhao, Erik Wijmans, Bhavana Jain, Julian Straub, Jia Liu, Vladlen Koltun, Jitendra Malik, et~al.
\newblock Habitat: A platform for embodied ai research.
\newblock In \emph{CVPR}, 2019.

\bibitem[Silberman et~al.(2012)Silberman, Hoiem, Kohli, and Fergus]{silberman2012indoor}
Nathan Silberman, Derek Hoiem, Pushmeet Kohli, and Rob Fergus.
\newblock Indoor segmentation and support inference from rgbd images.
\newblock In \emph{ECCV}, 2012.

\bibitem[Straub et~al.(2019)Straub, Whelan, Ma, Chen, Wijmans, Green, Engel, Mur-Artal, Ren, Verma, et~al.]{straub2019replica}
Julian Straub, Thomas Whelan, Lingni Ma, Yufan Chen, Erik Wijmans, Simon Green, Jakob~J Engel, Raul Mur-Artal, Carl Ren, Shobhit Verma, et~al.
\newblock The replica dataset: A digital replica of indoor spaces.
\newblock \emph{arXiv preprint arXiv:1906.05797}, 2019.

\bibitem[Tai et~al.(2018)Tai, Zhang, Liu, and Burgard]{tai2018socially}
Lei Tai, Jingwei Zhang, Ming Liu, and Wolfram Burgard.
\newblock Socially compliant navigation through raw depth inputs with generative adversarial imitation learning.
\newblock In \emph{ICRA}, 2018.

\bibitem[Vasiljevic et~al.(2019)Vasiljevic, Kolkin, Zhang, Luo, Wang, Dai, Daniele, Mostajabi, Basart, Walter, et~al.]{vasiljevic2019diode}
Igor Vasiljevic, Nick Kolkin, Shanyi Zhang, Ruotian Luo, Haochen Wang, Falcon~Z Dai, Andrea~F Daniele, Mohammadreza Mostajabi, Steven Basart, Matthew~R Walter, et~al.
\newblock Diode: A dense indoor and outdoor depth dataset.
\newblock \emph{arXiv preprint arXiv:1908.00463}, 2019.

\bibitem[Wu et~al.(2020)Wu, Hu, Happold, Xu, and Neumann]{wu2020geometry}
Cho-Ying Wu, Xiaoyan Hu, Michael Happold, Qiangeng Xu, and Ulrich Neumann.
\newblock Geometry-aware instance segmentation with disparity maps.
\newblock \emph{arXiv preprint arXiv:2006.07802}, 2020.

\bibitem[Wu et~al.(2022)Wu, Wang, Hall, Neumann, and Su]{wu2022toward}
Cho-Ying Wu, Jialiang Wang, Michael Hall, Ulrich Neumann, and Shuochen Su.
\newblock Toward practical monocular indoor depth estimation.
\newblock In \emph{CVPR}, 2022.

\bibitem[Wu et~al.(2023)Wu, Zhong, Wang, and Neumann]{wu2023meta}
Cho-Ying Wu, Yiqi Zhong, Junying Wang, and Ulrich Neumann.
\newblock Meta-optimization for higher model generalizability in single-image depth prediction.
\newblock \emph{arXiv preprint arXiv:2305.07269}, 2023.

\bibitem[Xie et~al.(2023)Xie, Geng, Hu, Zhang, Hu, and Cao]{xie2023revealing}
Zhenda Xie, Zigang Geng, Jingcheng Hu, Zheng Zhang, Han Hu, and Yue Cao.
\newblock Revealing the dark secrets of masked image modeling.
\newblock In \emph{CVPR}, 2023.

\bibitem[Yang et~al.(2024)Yang, Kang, Huang, Xu, Feng, and Zhao]{yang2024depth}
Lihe Yang, Bingyi Kang, Zilong Huang, Xiaogang Xu, Jiashi Feng, and Hengshuang Zhao.
\newblock Depth anything: Unleashing the power of large-scale unlabeled data.
\newblock In \emph{CVPR}, 2024.

\bibitem[Yin et~al.(2019)Yin, Liu, Shen, and Yan]{yin2019enforcing}
Wei Yin, Yifan Liu, Chunhua Shen, and Youliang Yan.
\newblock Enforcing geometric constraints of virtual normal for depth prediction.
\newblock In \emph{ICCV}, 2019.

\bibitem[Yin et~al.(2021)Yin, Zhang, Wang, Niklaus, Mai, Chen, and Shen]{yin2021learning}
Wei Yin, Jianming Zhang, Oliver Wang, Simon Niklaus, Long Mai, Simon Chen, and Chunhua Shen.
\newblock Learning to recover 3d scene shape from a single image.
\newblock In \emph{CVPR}, 2021.

\bibitem[Yuan et~al.(2022)Yuan, Gu, Dai, Zhu, and Tan]{yuan2022new}
Weihao Yuan, Xiaodong Gu, Zuozhuo Dai, Siyu Zhu, and Ping Tan.
\newblock New crfs: Neural window fully-connected crfs for monocular depth estimation.
\newblock \emph{CVPR}, 2022.

\bibitem[Zhao et~al.(2023{\natexlab{a}})Zhao, Poggi, Tosi, Zhou, Sun, Tang, and Mattoccia]{Zhao_2023_ICCV}
Chaoqiang Zhao, Matteo Poggi, Fabio Tosi, Lei Zhou, Qiyu Sun, Yang Tang, and Stefano Mattoccia.
\newblock Gasmono: Geometry-aided self-supervised monocular depth estimation for indoor scenes.
\newblock In \emph{ICCV}, 2023{\natexlab{a}}.

\bibitem[Zhao et~al.(2023{\natexlab{b}})Zhao, Rao, Liu, Liu, Zhou, and Lu]{zhao2023unleashing}
Wenliang Zhao, Yongming Rao, Zuyan Liu, Benlin Liu, Jie Zhou, and Jiwen Lu.
\newblock Unleashing text-to-image diffusion models for visual perception.
\newblock \emph{ICCV}, 2023{\natexlab{b}}.

\end{thebibliography}
}
\end{document}